**Full title:** Cardiac CT perfusion imaging of pericoronary adipose tissue (PCAT) highlights potential confounds in coronary CTA

Hao Wu,[a] BS; Yingnan Song,[a] BS; Ammar Hoori,[a] PhD; Ananya Subramaniam,[a] BS; Juhwan Lee,[a] PhD; Justin Kim,[a] BS; Tao Hu,[a] BS; Sadeer Al-Kindi,[b] MD; Wei-Ming Huang,[c] MD; Chun-Ho Yun,[c] MD; Chung-Lieh Hung,[d] MD; Sanjay Rajagopalan,[b,e] MD; David L. Wilson,[a,f] PhD

[a] Department of Biomedical Engineering, Case Western Reserve University, Cleveland, OH, 44106, USA
[b] Harrington Heart and Vascular Institute, University Hospitals Cleveland Medical center, Cleveland, OH, 44106, USA
[c] Department of Radiology, MacKay Memorial Hospital, Taipei, Taiwan
[d] Division of Cardiology, Department of Internal Medicine, MacKay Memorial Hospital, Taipei, Taiwan
[e] School of Medicine, Case Western Reserve University, Cleveland, OH, 44106, USA
[f] Department of Radiology, Case Western Reserve University, Cleveland, OH, 44106, USA

**Corresponding author**
Sanjay Rajagopalan
Harrington Heart and Vascular Institute, University Hospitals Cleveland Medical center, Cleveland, OH, 44106,
Sanjay.Rajagopalan@UHhospitals.org
734-646-6758

David L Wilson
Department of Biomedical Engineering, Case Western Reserve University, 10900 Euclid Avenue, Cleveland OH 44106
david.wilson@case.edu
216-368-4099

**Declaration.** Human subject research has been done under an IRB of Case Western Reserve University and University Hospitals of Cleveland. Images were acquired at Mackay memorial hospital, Taiwan, and shared under a data use agreement. Case Western Reserve University has licensed the CCTP analysis technology to BioInVision, Inc.; David L. Wilson is part owner of BioInVision, Inc. This information has been disclosed to Case Western Reserve University, and David L. Wilson has an approved CWRU plan for managing potential conflicts of interest. This research was supported by National Heart, Lung, and Blood Institute through grants R01HL167199, R01HL165218, R01 HL143484, and R44HL156811. The grants were obtained via collaboration between Case Western Reserve University and University Hospitals of Cleveland.

## Abstract

**Background.** Features of pericoronary adipose tissue (PCAT) assessed from coronary computed tomography angiography (CCTA) are associated with inflammation and cardiovascular risk. As PCAT is vascularly connected with coronary vasculature, the presence of iodine is a potential confounding factor on PCAT HU and textures that has not been adequately investigated.

**Objective.** Use dynamic cardiac CT perfusion (CCTP) to inform contrast determinants of PCAT assessment.

**Method.** From CCTP, we analyzed HU dynamics of territory-specific PCAT, myocardium, and other adipose depots in patients with coronary artery disease. HU, blood flow, and radiomics were assessed over time. Changes from peak aorta time, Pa, chosen to model the time of CCTA, were obtained.

**Results**. HU in PCAT increased more than in other adipose depots. The estimated blood flow in PCAT was ~23% of that in the contiguous myocardium. Comparing PCAT distal and proximal to a significant stenosis, we found less enhancement and longer time-to-peak distally. Two-second offsets [before, after] Pa resulted in [-4±1.1-HU, 3±1.5-HU] differences in PCAT. Due to changes in HU, the apparent PCAT volume reduced ~15% from the first scan (P1) to Pa using a conventional fat window. Comparing radiomic features over time, 78% of features changed >10% relative to P1.

**Conclusion.** CCTP elucidates blood flow in PCAT and enables analysis of PCAT features over time. PCAT assessments (HU, apparent volume, and radiomics) are sensitive to acquisition timing and the presence of obstructive stenosis, which may confound the interpretation of PCAT in CCTA images. Data normalization may be in order.

**Abbreviations List:**

CCTP = cardiac CT perfusion.
CCTA = coronary computed tomography angiography.
PCAT = pericoronary adipose tissue.
SUB = subcutaneous adipose tissue.
PAT = paracardial adipose tissue.
PAAT = periaortic adipose tissue.
EAT = epicardial adipose tissue.
MYO = myocardium.
Pa = aorta peak enhancement time point.
Ppcat = of PCAT peak enhancement time point.

**Introduction**

Pericoronary adipose tissue (PCAT) assessment by coronary CT angiography (CCTA) has emerged as a novel CT-imaging-based biomarker associated with plaque vulnerability, plaque-specific inflammation, and cardiovascular risk (1–3). Fat attenuation index (FAI), related to the regional average HU value, has been reported to be higher in lesions with higher stenosis severity and has been associated with vascular inflammation as proven from biopsy samples taken from patients undergoing cardiac surgery (4). In addition, others have reported higher PCAT HU in lesions with higher stenosis severity, in lesions with higher rates of atherosclerosis progression, and in culprit vs non-culprit lesions in patients presenting with acute coronary syndromes (1,5). CCTA-based radiomics profiling of coronary artery PCAT detects perivascular structural remodeling associated with coronary artery disease and improve cardiovascular risk prediction (6,7).

These previous studies were all performed in the presence of iodine contrast which could be a potential confounding factor in the assessment of PCAT. Although the blood flow in adipose is less than in other tissues, it is expected that perfusion of iodine will elevate adipose HU values with effects on other characteristics such as apparent volume and texture radiomics. Indeed a recent study from Almeida et al., found that iodine contrast may affect the apparent volume of PCAT and demonstrated the feasibility of assessing PCAT from non-contrast images (8). They found that before contrast injection, the average HU value was lower and volume was increased when using the traditional fat window (-190-HU, -30-HU). Continuing with this line of reasoning, we believe that there could be a dependence of HU, volume, and texture radiomics features upon the time since injection as the iodine contrast bolus moves through the vasculature and upon the presence of a flow-limiting stenosis. In addition, coronary artery motion, scan parameters (i.e., kVp), beam hardening effects from contrast agent, scanner type, scanner calibration, and body size, may all potentially influence PCAT assessment.

In this report, we use dynamic CCTP to elucidate the role of iodine dynamics on PCAT assessment. We measure PCAT blood flow and determine changes in HU values of PCAT, in comparison to the myocardium, and other adipose depots. Using images at the time of peak contrast in the aorta (Pa), to model a CCTA acquisition, we compared HU and radiomics assessment to determine the potential confound due to variable acquisition timing. As obstructed vessels will have reduced flows, we also compared PCAT with and without obstructive disease. Our goal is to elucidate the extent to which contrast enhancement with CCTA leads to uncertainty in conventional PCAT assessments.

**Methods**

This study was approved as a retrospective study of de-identified data by the local institutional review board. Images were acquired starting in 2013 at Mackay memorial hospital, Taiwan, and shared under a data use agreement. We analyzed 9 patients with suspected coronary artery disease (CAD) who underwent CCTA and stress CCTP. We excluded patients based on the following criteria: 1) age<20 years, 2) coronary artery bypass grafting, 3) acute or old myocardial infarction, 4) complete left bundle branch block, 5) inadequate datasets such as poor image quality of CCTA, insufficient CCTP analysis. Sequential CCTA, and stress dynamic CCTP were both performed using a dual-source CT system (Somatom Definition Flash; Siemens Healthineer, Forchheim, Germany) with a collimation of 64x2x0.6mm and flying-focal spot, resulting in 2x128 sections. Details of CCTA and CCTP image acquisition can be found in supplemental material.

The CCTA data were evaluated by two experienced readers (CHY, WMH, with 16 and 6 years of cardiac CT experience, respectively). Readers were blinded to the subject's clinical presentation and history. Any disagreement was solved by a consensus. On a per vessel basis, stenoses were classified as: normal, 0% luminal diameter stenosis; minimal, 1% to 24% stenosis; mild, 25% to 49% stenosis; moderate, 50% to 69% stenosis; severe, 70% to 99% stenosis; and occluded, 100% stenosis. Stenosis >70% were considered as obstructive stenosis.

We used in-house CCTP analysis software for CCTP image processing (see central illustration figure) as described in our previous publications (9–11). Briefly, we performed a calibration-free beam-hardening correction for each scan (9). To reduce the motion artifacts between scans, we performed a two-step registration (rigid body followed by non-rigid) to register all scans to the scan with peak enhancement of aorta as described previously (12). We applied a spatio-temporal bilateral filter to the registered scans to further reduce noise. Myocardium (MYO) and aorta were segmented automatically using U-Net (13). ROIs for paracardial adipose tissue (PAT), epicardial adipose tissue (EAT), periaortic adipose tissue (PAAT), and subcutaneous adipose tissue (SUB) were manually segmented. To segment PCAT, the coronary lumen (LAD and RCA) was segmented manually by using the aorta peak enhancement (Pa) scan. We then created binary masks in axial images (deemed axial-disk-masks), which were centered on the centerline with a diameter 2-times the median effective diameter of the first 40-mm segmented lumen of coronaries. Finally, voxels in the axial-disk-mask within the fat window (-190 HU, -30 HU) were selected as PCAT. Blood flow were computed for myocardium and PCAT region based on a super-voxel based robust physiologic model (10,11).

PCAT features were extracted from voxels within the binary masks at each time point. We extracted both hand-crafted (13 features) and radiomics (8 features) from PCAT region. For hand-crafted features, we focused on the histogram feature of PCAT HU values (e.g. skewness, kurtosis of the HU value histogram, and the probability in range of HU values) and the PCAT axial area features (e.g., mean area of PCAT in axial view). For the radiomics feature, we extract features that sensitive to predict coronary artery disease reported by other studies via PyRadiomics library (14–16). We included shape features (e.g. original-shape-Flatness and original-shape-Elongation) and wavelet features (e.g., wavelet-LLH-firstorder-Mean and wavelet-LLL-glcm-Idmn). Texture features were calculated using PCAT voxels with 16 bins discretization. Radiomic features were extracted from both original PCAT images and three-dimensional wavelet transformations of the original image. Wavelet transformation decomposes the data into high and low-frequency components, enabling capturing discontinuities, ruptures and singularities and coarse structure of the data.

We statistically analyzed features. ROIs include PCAT, subcutaneous adipose tissue (SUB) paracardial adipose tissue (PAT), epicardial adipose tissue (EAT), periaortic adipose tissue (PAAT), myocardium (MYO), and aorta. We examined temporal changes in attenuation and compared PCAT attenuation, apparent volume and radiomics between time points. Some special time points are: 1. P1, the first time point which is the pre-contrast time point. 2. Pa, the peak enhancement time point of aorta (Pa) which is equivalent to the CCTA acquisition time point. 3. Ppcat, the peak enhancement time point of PCAT. Each measurement was expressed as mean ± standard error. The unpaired Student's t-test was used to compare the means between two groups. Differences were deemed statistically significant at $P < 0.05$. We used MATLAB statistic toolbox (17).

**Results**

Iodine enhancement was analyzed in various adipose depots for a patient with adequate myocardial perfusion in the presence of unobstructed coronary arteries (Fig. 1). We found that adipose tissue in general, show much less enhancement than the myocardium. PCAT and EAT, however enhanced much more than other fat depots, and followed a time course similar to that of the myocardium. In contrast other adipose tissues (PAT, SUB, and PAAT) had relatively flat attenuation curves. PCAT enhanced over time, giving a peak change of ~22-HU (P1: -75±5-HU, Ppcat: -53±6-HU, p<0.05). We compared PCAT enhancement to that of ROIs in the LAD and myocardium (Fig. 2.). The HU time curve for PCAT was remarkably similar to that for the nearby myocardium. For this patient without obstructive coronary artery disease, the mean blood flow in PCAT was 23% of that in the myocardium (75 vs. 324 mL/100g-min).

<<insert figure 1>>

<<insert figure 2>>

We examined HU values in adipose tissue depots and determined the amount of variation that may be expected when sampling at different time points. In Fig. 3, show that baseline HU values at P1 were different for different fat depots, with the highest value in PAAT, which was 24±3-HU higher than that for PAT, indicating substantially different tissue characteristics. The change of HU in EAT was greater than for other types of adipose tissue. In Fig. 4, across 9 patients, we analyzed ΔHU values relative to P1, providing some correction for individual differences. Over all scans, ΔHU values were under 5 HU for fat depots SUB, PAT, and EAT. PCAT ΔHU values relative to P1, for both LAD and RCA, far exceeded those in other depots. In addition, changes relative to Pa may exceed 8-HU, underscoring that the timing of a CCTA acquisition may affect average HU values on PCAT. In addition, for PCAT around vessels without obstructive stenosis, the HU values were -74±1.9, -70±2.1, and -67±2.9, at Pa-1, Pa, and Pa+1, respectively, given -4±1.1 to 3±1.5-HU difference with two seconds offset (before and after) in acquisition around Pa, Fig. 5(c).

<<insert figure 3>>

<<insert figure 4>>

We analyzed EAT at varying distances from coronary arteries. In Fig. 5, we analyzed "remote" EAT and compared results to all EAT and PCAT. Baseline values were over 9-HU higher for PCAT than remote EAT, suggesting substantially different characteristics (Fig. 5(c)). In addition, PCAT demonstrated a greater change (~12-HU) with iodine enhancement, while remote EAT changes only about 3-HU, suggesting that PCAT is more substantively perfused. All EAT, including PCAT, shows a similar trend to PCAT, but a smaller change. In addition, we analyzed enhancement of PCAT at different radial distances from the vessel wall to try to determine a potential rationale for the normal PCAT region definition of two times the diameter of the vessel wall. In Fig. S1, we analyzed different disk and annular regions as defined in the figure. Baseline HU values were higher for the inner disk than the two annular rings. In addition, ΔHU from P1 to Ppcat progressively decreased as one goes out from the vessel wall.

<<insert figure 5>>

Given the known relationship between reduced myocardial blood flow in territories distal to a significant obstructive stenosis, we wanted to determine if there were similar effects on PCAT. In Fig. 6, we analyzed HU values in PCAT regions proximal and distal to an obstructive lesion. Distal to the lesion, there was a delay and diminution of the peak, for both myocardium and PCAT as compared to curves obtained proximal to the lesion. The diminution in peaks were present even when we evaluated ΔHU by subtracting baseline values. We found that four out of six vessels with severe stenosis followed a similar trend. The ΔHU between P1 and the peak enhancement for PCAT were 16.1±6.7-HU and 12.3±3.4-HU proximal and distal of a lesion, respectively.

<<insert figure 6>>

In addition to effect on HU, we investigated the role of iodine perfusion on other features of interest such as apparent PCAT volume (volume of voxels within the axial-radial-disks falling within the standard fat window [-190-HU, -30-HU]). Using this definition, the apparent PCAT volume changed substantially over time (Fig. 7). As voxels were "lost" due to their enhancement with iodine, we investigated an extended fat window [-190-HU, -10-HU]. In Fig. 7(a), the apparent PCAT volume at LAD for a patient changed from 3.4 cm$^3$ to 2.9 cm$^3$ from P1 to Pa, giving a 14.7% reduction using the standard fat window. Using the extended fat window, the change was 12.3% (not shown). In Fig. 7(b), over all vessels (n = 18), the apparent PCAT volume at Pa changed by means of ~15% and ~11% relative to the volume at P1, for the standard and the extended fat window, respectively.

<<insert figure 7>>

In addition to volume, we considered that the presence of iodine could change other radiomic features, including texture which would likely depend upon vessel filling. In Fig. 8, we analyzed 13 hand-crafted features and the 8 radiomic features as suggested in the literature (14,15) over time. Only 5 (22%) of hand-crafted and radiomics features changed values less than 10% relative to P1. They are: entropy, original-shape-Elongation, original-shape-Flatness, wavelet-LHL-firstorder-Kurtosis, and wavelet-LLL-glcm-ldmn. Other features such as area mean change over time in a manner similar to volume, as described previously. Note that a change of ±2 scan intervals from Pa can dramatically change features values suggesting a dependence on the CCTA acquisition time, e.g, see wavelet-HHH-glszm-SizeZoneNonUniformityNormalized.

<<insert figure 8>>

**Discussion**

In this work we have provided important new information on the influence of iodine contrast on PCAT assessment. The use of cardiac CT perfusion (CCTP) provides substantial clarity on the influence of contrast kinetics on PCAT assessment in a number of ways. First, PCAT substantively enhances with iodine in concert with the myocardium, with a blood flow of about 1/5 of that of the myocardium. Other adipose depots (e.g., paracardial, subcutaneous, and periaortic) show little, if any enhancement. (This could be due to both reduced vascularization as compared to PCAT and to spreading of the iodine bolus in feeding arteries.) Together, these observations suggest that unlike other depots, PCAT and EAT are in vascular communication with myocardial and coronary tissues. Second, dynamic PCAT enhancement with iodine, confounds HU, apparent volume, and radiomic features, depending on the timing of the CCTA acquisition. Third, interpretation of PCAT assessment may be further influenced by the presence of obstructive coronary disease.

Our results also provide new information on the vascularity of PCAT, when contrasted with other adipose depots. For instance, we found that the HU time course in EAT, demonstrated considerably more enhancement than other types of adipose tissue, outside the pericardium (Figs. 1, 3, and 4). This finding is consistent with results in the literature suggesting that blood flow in visceral adipose tissue far exceeds that of subcutaneous adipose tissue (18), perhaps related to the underlying metabolic requirements of the visceral organs such as the heart. The EAT would be expected to demonstrate higher perfusion, given the underlying metabolic requirements of the heart contiguous to it with a shared vascular supply. Imaging of the vasa vasorum shows dense vascularization (19) much of which would be in contact with surrounding tissues including adipose (20), supporting our observation of increased enhancement of PCAT near the coronaries that drops rapidly within two vessel diameters distance (Fig. S1). Given the metabolic requirements of the coronary arteries (which in turn feed the myocardium), the importance of subserving flow to the coronary arteries cannot be overstated. PCAT has been noted to have a different transcriptomic signature than that of other EAT regions (21).

The timing of a CCTA acquisition can also affect multiple assessments including HU values, apparent volume, and radiomics which has important implications for studies that report these characteristics without consideration of contrast kinetics. A two-second offset [before, and after] in acquisition around Pa results in differences [-4±1.1 to 3±1.5-HU difference] in PCAT attenuation. We observed up to a ~15% apparent volume difference between the pre-contrast scan and the peak time point, with the standard fat window [-190-HU -30-HU]. A reduced change in apparent volume (~ 11%) was observed when using the adjusted window [-190 -10] HU, indicating that some PCAT voxels were enhanced by iodine and exceed -30 HU. The HU and volume changes could further affect the intensity-based and shape-based features when computing radiomics of PCAT in CCTA. We computed 21 hand-crafted and radiomics features in PCAT at different time points. Only 22% of hand-crafted and radiomics features changed less than 10% relative to the pre-contrast scan, suggesting a large influence of iodine in PCAT radiomics analysis in CCTA studies.

The presence of a proximal obstructive coronary artery stenosis could further impact PCAT attenuation. In general, we found that HU values of PCAT were lower and the time to peak enhancement were longer in PCAT regions distal to the stenosis, consistent with reduced flow in the region and reduced perfusion of the distal subtended myocardium. In earlier reports PCAT HU was higher in the presence of CAD, defined as >50% stenosis, with increases of ~3-HU (4) and ~5-HU (22). These studies did not interpret these values with attention to consideration of contrast timing and therefore are difficult to interpret. In more recent reports, authors have examined more severely obstructed vessels. Ma et al reported that the lesion specific PCAT mean HU value was ~4 HU lower in vessels with >70% stenosis than in vessel with <25% stenosis (23). They argued that there might be more stable calcification and less inflammation in the presence of a severe stenosis as compared to mild and moderate stenoses. While these conjectured scenarios seem attractive from a biologic perspective, it is hard to support without rigorous adjustments and consideration of other plausible hemodynamic considerations. For instance, the pressure drop across an obstructive stenosis may reduce flow to the distal region, effectively reducing microcirculatory flow to the coronary microvessels and coronary vasa vasorum distal the stenosis. This is indeed evidenced by our results demonstrating a reduction in distal myocardial perfusion and reduced enhancement in the distal PCAT. In the aforementioned CCTA studies, it is unclear to what extent some of the observed differences therefore relate to hemodynamic considerations.

In addition to identifying potential confounding factors, our analysis with CCTP might lead to improvements in assessment of PCAT. As vessel density (vascularity) and vessel leakiness (reflective of neo-angiogenesis) might change with inflammation (24), perfusion metrics (e.g., blood flow and retention) might be important features for risk prediction and may indeed capture some of these features. In addition, additional analysis of CCTP data might lead to ways to normalize responses (e.g., normalizing on aorta or vessel HU values) for improved characterization.

We admit several limitations including the single-center nature of this study with a predominantly Asian patient population. However, there is no reason to believe that these aspects would detract from the broader relevance of our findings. Additionally, since CCTA was done before the CCTP in this study, the residual iodine from the CCTA study may have elevated baseline HU values of adipose. The performance of CCTP after a 10-mins delay of CCTA for washout of contrast medium, as recommended (25), helped reduce the impact of residual iodine. Further, ΔHU values should not be affected by the presence of residual iodine. Additional technical considerations include the choice of timing parameters for CCTA acquisition, should not discredit our observations. Finally, our CCTP acquisition which differ from CCTA imaging with regards to kVp, iodine bolus, acquisition method (e.g., shuttle versus spiral), and reconstruction are issues worth considering. However, given the fact that the CCTP methodology allows elucidation of the influence of a number of factors critical to the interpretation of PCAT attenuation and radiomics, our study findings have direct relevance to the field.

**Conclusion**

Our findings provide new understanding on PCAT perfusion and its assessment in CCTA images, which may need to contextualize factors such as acquisition timing and the presence of significant obstructive coronary stenosis, which may all confound PCAT HU values and other feature assessments.

**Clinical perspectives**

**Competency in medical knowledge:** The use of cardiac CT perfusion (CCTP) clarifies the influence of contrast kinetics on PCAT assessment. Observations suggest that PCAT and EAT are in vascular communication with myocardial and coronary tissues. Dynamic PCAT enhancement with iodine can make HU, apparent volume, and radiomic features, dependent on CCTA acquisition timing. Interpretation of PCAT assessment may be further influenced by the presence of obstructive coronary disease.

**Translational outlook:** Regularization of CCTA acquisitions in PCAT analysis are needed and image normalizations might be in order.

**Acknowledgments:**

This research was supported by National Heart, Lung, and Blood Institute through grants R01HL167199, R01HL165218, R01 HL143484, and R44HL156811. The content of this report is solely the responsibility of the authors and does not necessarily represent the official views of the National Institutes of Health. The grants were obtained via collaboration between Case Western Reserve University and University Hospitals Cleveland Medical Center. The veracity guarantor, Yingnan Song, affirms to the best of his knowledge that all aspects of this paper are accurate.

**Figure titles and legends**

Figure 1. HU time courses in ROIs of a patient with unobstructed coronary arteries and adequate myocardial perfusion. Color-coded plots in (b) correspond to the ROIs identified in (a)-bottom. In (a)-top, HU values are shown in PCAT around the LAD at different scan times. HUs are elevated at scans-7 and -11 compared to scan-1. There are 11 scans comprising ~22 seconds in a shuttle mode acquisition. The vertical dashed line shows the scan of peak enhancement in the aorta, which we deem the CCTA imaging time point, Pa. EAT = epicardial adipose tissue; HU = Hounsfield unit; MYO = myocardium; PAAT = periaortic adipose tissue; PAT = paracardial adipose tissue; PCAT = pericoronary adipose tissue; ROI = region of interest; SUB = subcutaneous adipose tissue.

Figure 2. Comparison of PCAT enhancement to LAD and nearby myocardium. CCTP image shows iodine in PCAT of LAD and myocardium (a). (See Video.1 for dynamic iodine enhancement.) Curved-planar-reformatted images of LAD for scans 1, 5, and 9 are shown in (b), (c) and (d), respectively. Note the enhancement in the middle scan. Enhancement time courses of LAD, PCAT, and nearby myocardium are remarkably similar in (e), (f), and (g), respectively. We computed the blood flow of myocardial and PCAT around LAD in (h) using our software described in Methods. Abbreviations as in figure 1.

Figure 3. HUs of adipose ROIs at different times. HUs at the first scan (P1) vary greatly between ROIs with the lowest value in pericardial adipose (PAT) and the highest in periaortic adipose (PAAT). Other times are relative to the time of peak enhancement of aorta (Pa). EAT changes more than other ROIs. Abbreviations as in figure 1.

Figure 4. HU changes relative to baseline (P1) for various adipose depots and time points across 9 patients. ΔHU values order PCAT> EAT>PAT≈SUB. PCAT is similar in LAD and RCA.

Maximum enhancement (at Ppcat), is significantly different from that at Pa, suggesting enhancement delay in some hearts. Average time for Ppcat is Pa+2. Only vessels without obstructive disease are used (6-LADs and 6-RCAs). *P<0.05. Abbreviations as in figure 1.

Figure 5. Regional HU values within EAT (containing PCAT and remote EAT). EAT and remote EAT ROIs are shown in (a) and (b), respectively. Baseline HU and change with enhancement vary across ROIs, with PCAT changing the most (c). Only vessels without obstructive disease are used (6-LADs and 6-RCAs). Abbreviations as in figure 1.

Figure 6. HU time courses in ROIs of a patient with an obstructive lesion and reduced myocardial blood flow in the RCA territory. Severe stenosis on the RCA (a) corresponds to reduced proximal myocardial blood flow (b). Distal as compared to proximal PCAT ROIs [blue and brown regions in (a)] gives a delayed and depressed enhancement time course (c-bottom), which are reasonably mirrored in the myocardial regions. Perfusion pressure proximal to the lesion will be increased relative to that distal.

Figure 7. PCAT apparent volume changes over time. In (a), LAD PCAT apparent volume change (bottom) is shown for the patient (above) with non-obstructive disease and fat threshold [-190-HU, -30-HU]. In (b), the percentage of apparent volume change relative to P1 is shown across all vessels with the conventional threshold and an adjusted threshold [-190-HU, -10-HU]. The solid line is the mean, and the shadow is the standard error. Percent apparent volume changes are substantial but reduced with the adjusted threshold.

Figure 8. Changes in feature values with iodine perfusion. Twenty-one hand-crafted and radiomic library features are on the vertical axis. We present mean percent changes as compared to values

at P1 from Pa-4 to Pa+4. The range of change was [-214%, 460%] but data were truncated to [-30%, 30%] for visualization. We analyzed vessel territories without obstruction (n = 12).

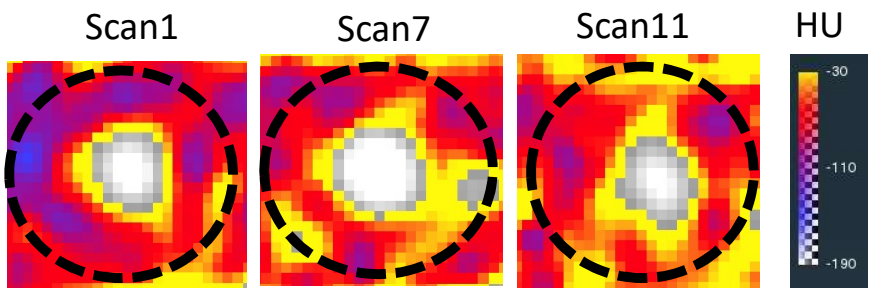
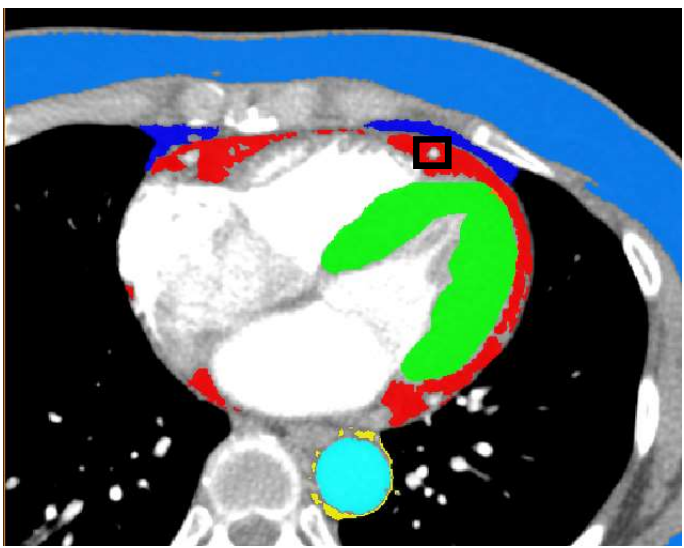
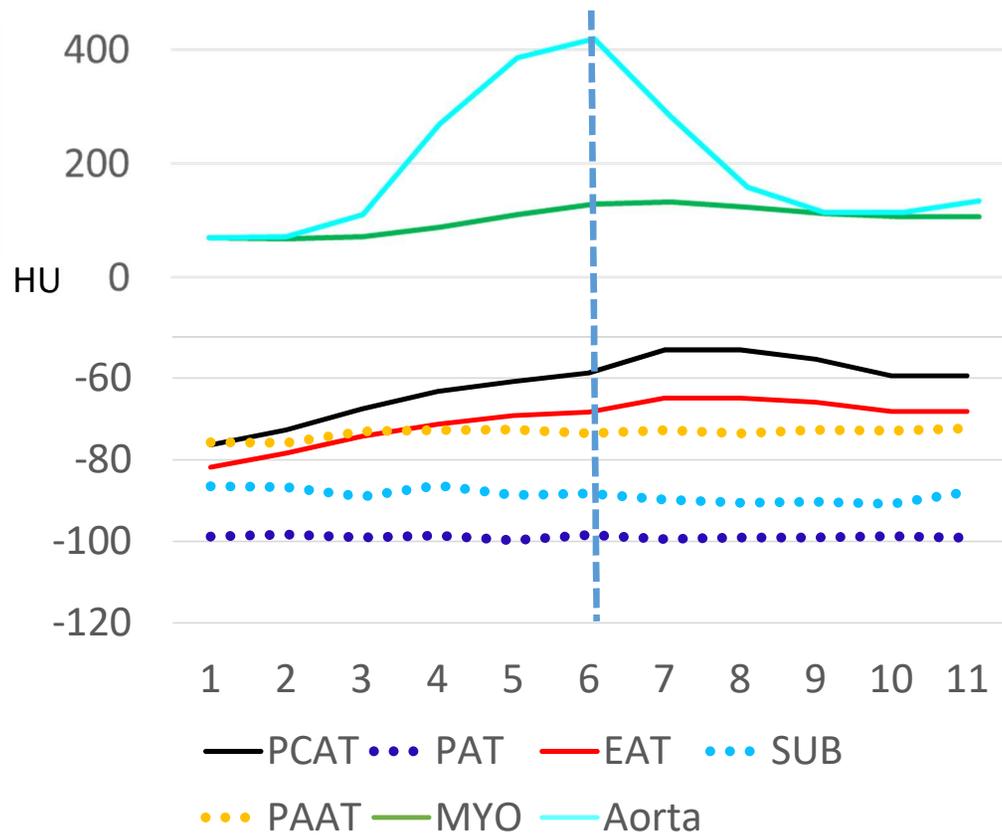

(a) (b)

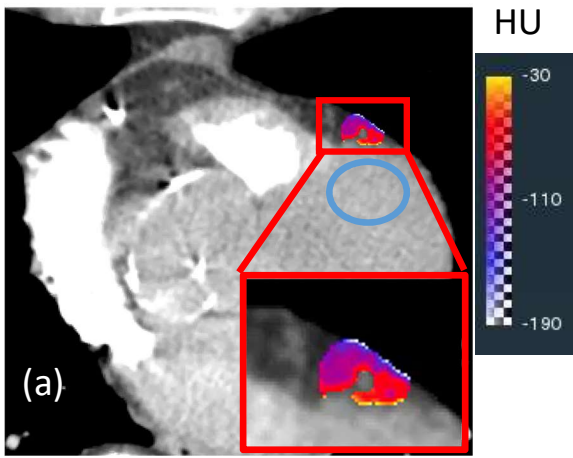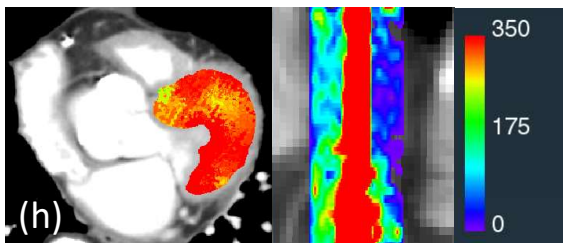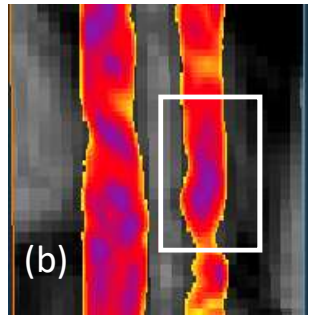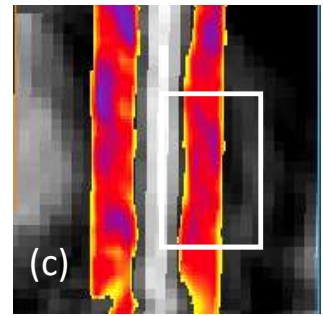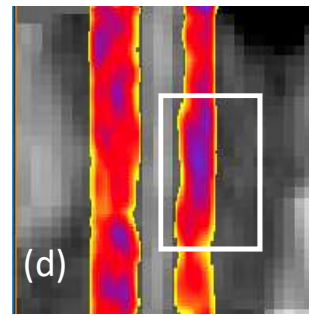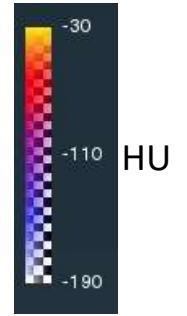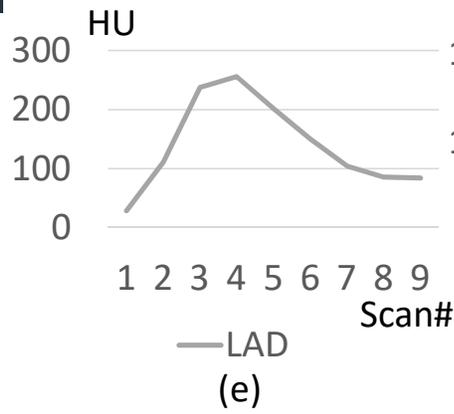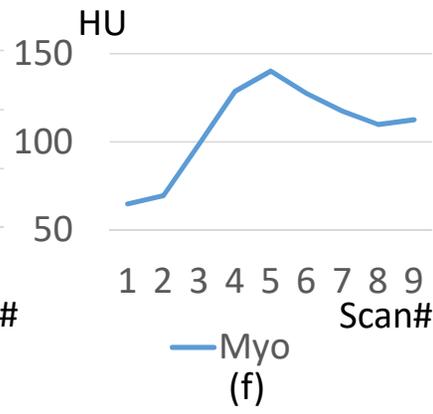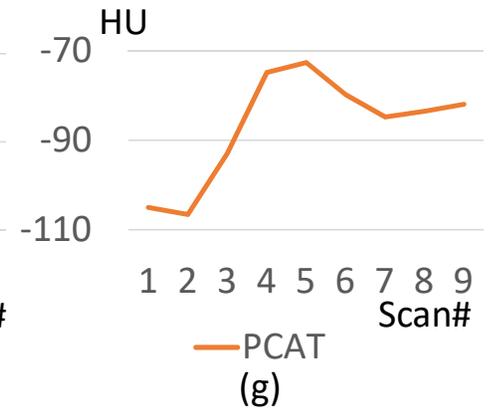

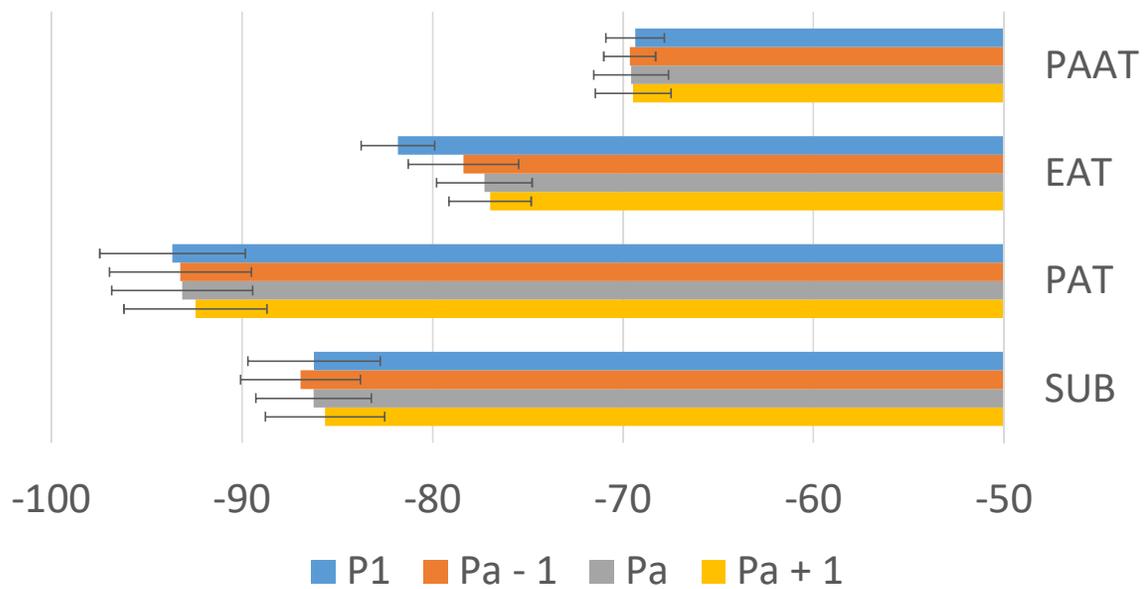

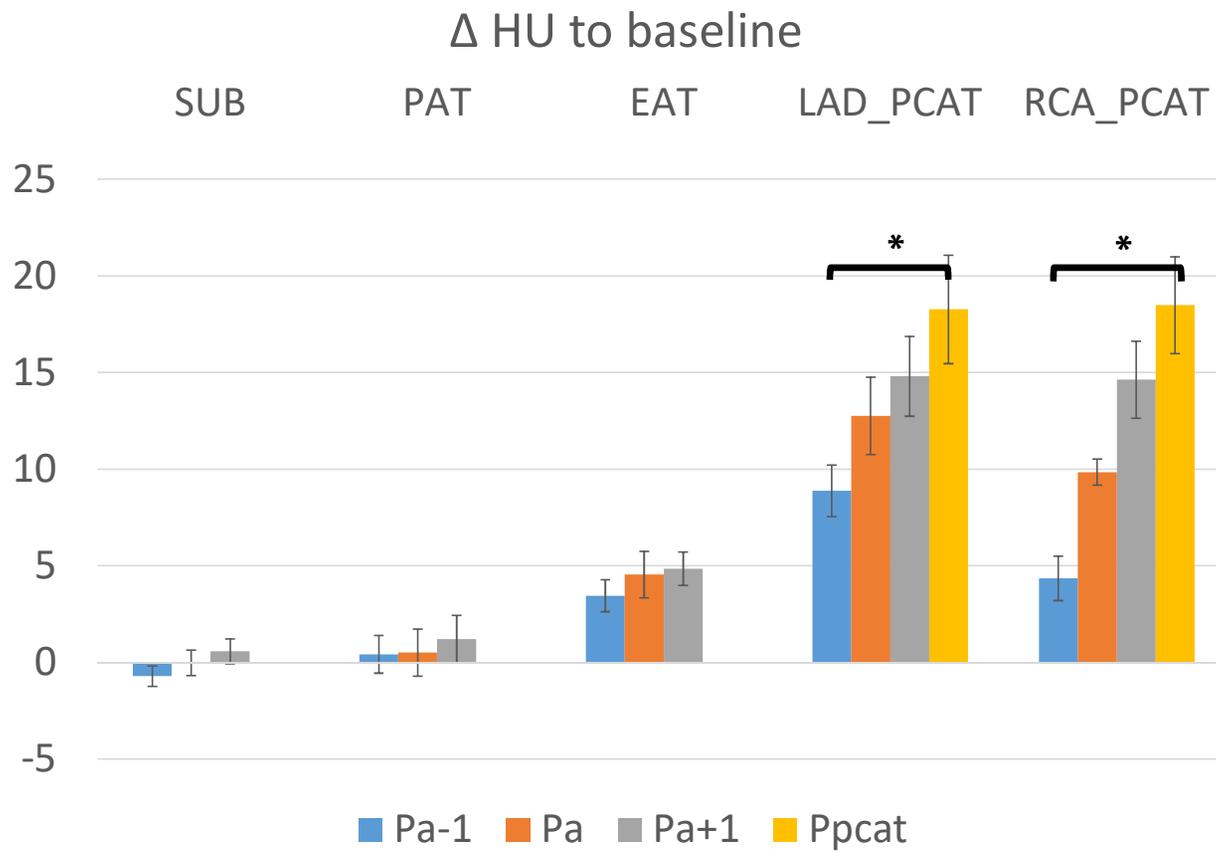

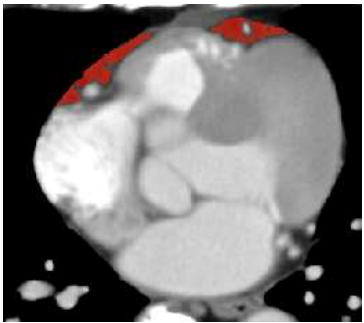
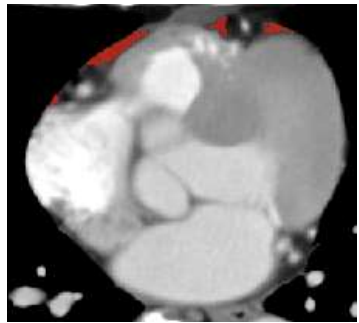
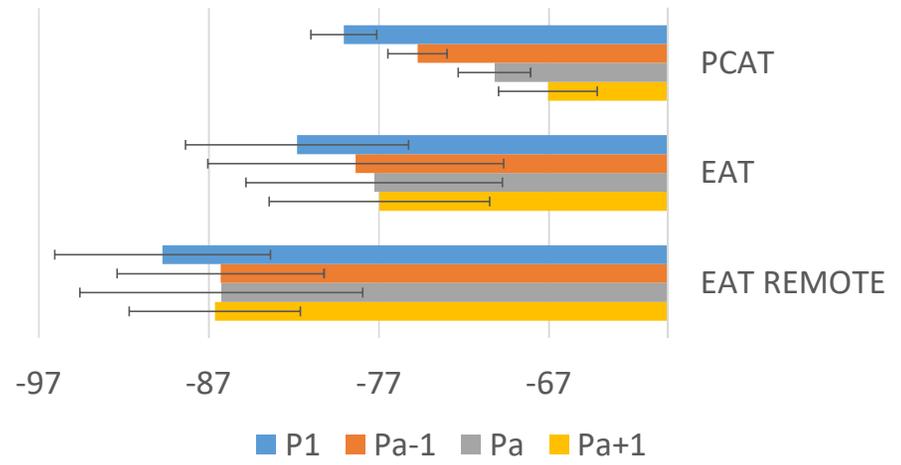

(a)          (b)          (c)

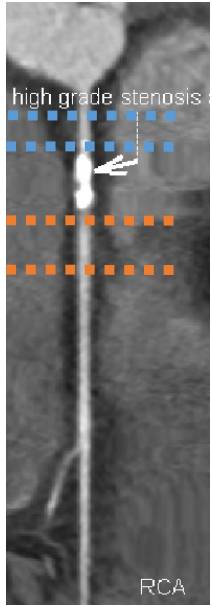 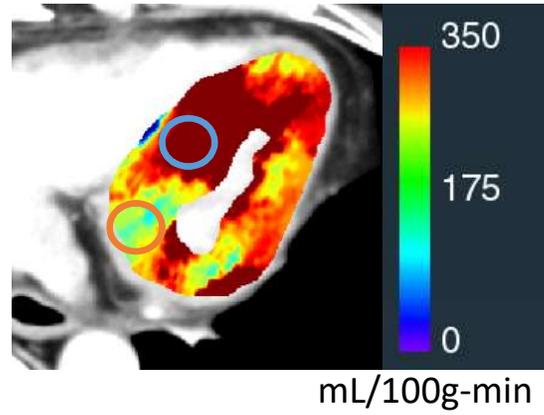 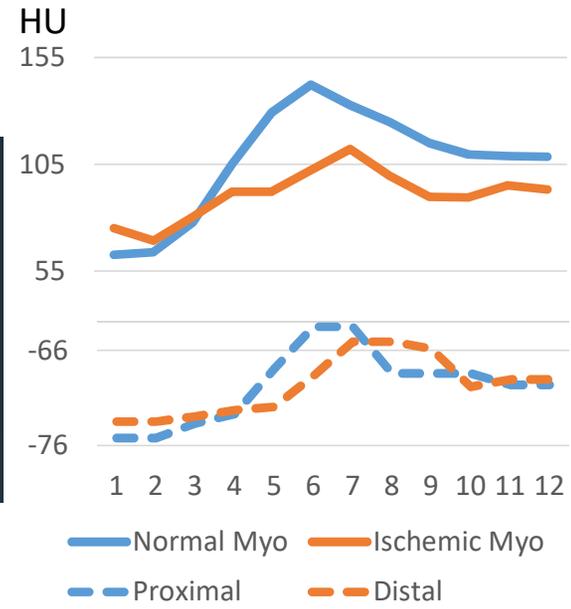

(a) (b) (c)

Scan 1  Scan 5  Scan 9

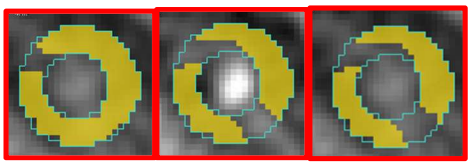

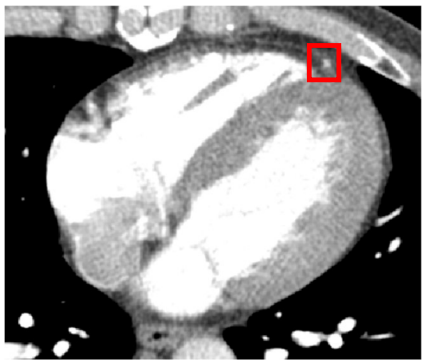

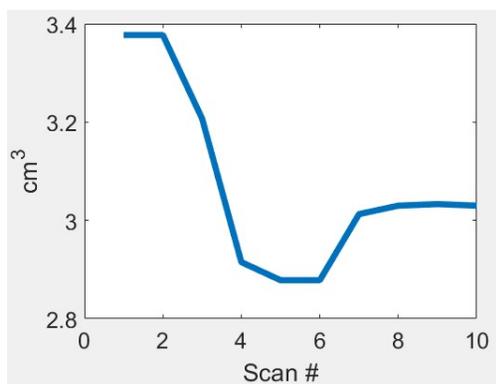

(a)

Δ% volume of PCAT across all vessels (n = 18)

Standard window: [-190 -30]

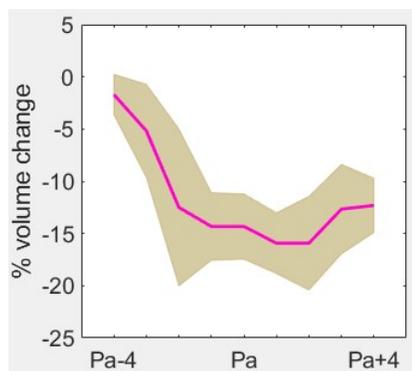

Adjusted window : [-190 -10]

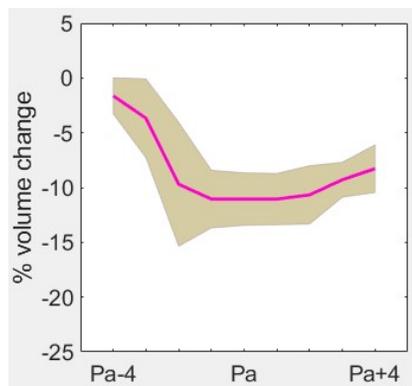

(b)

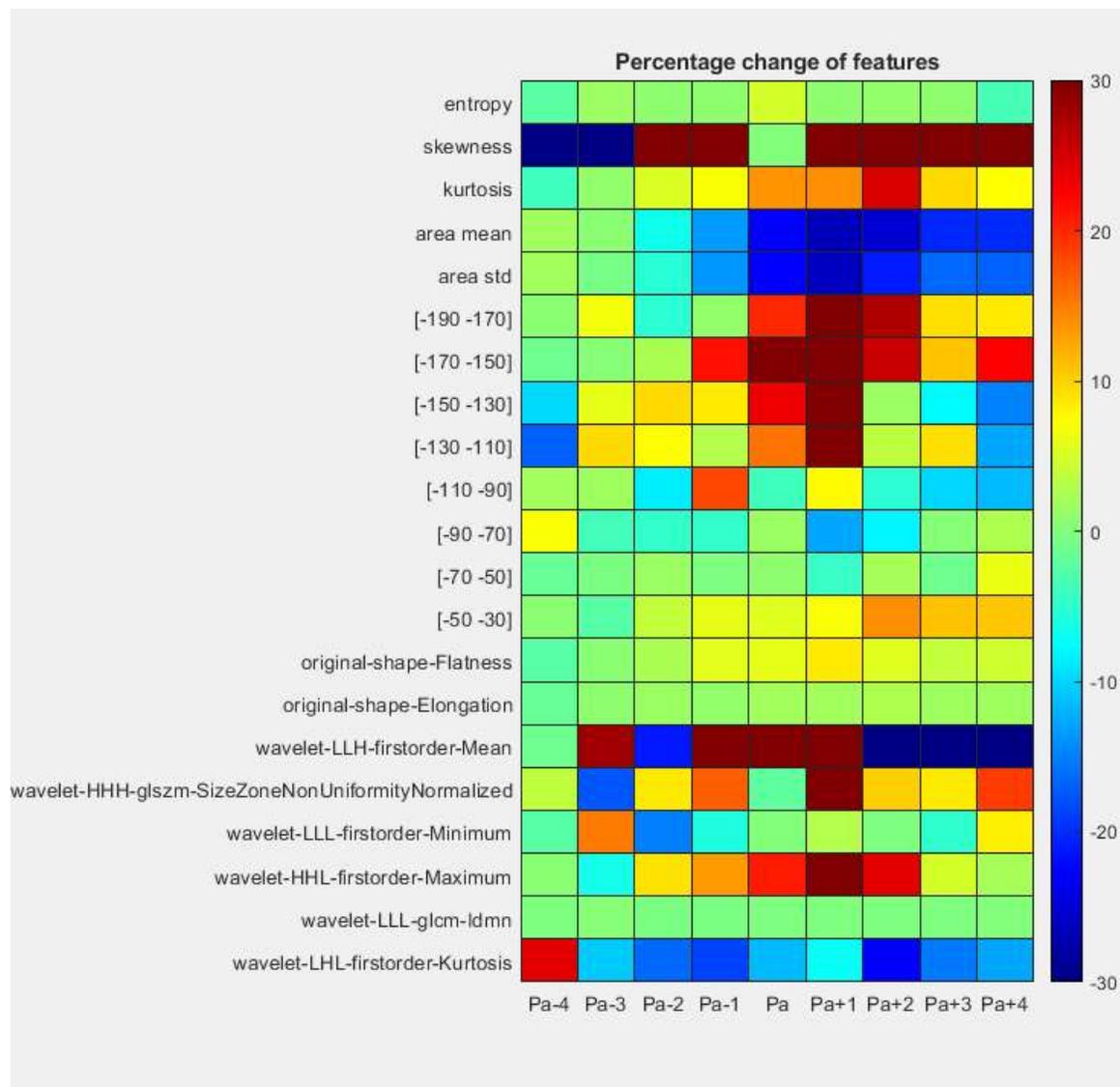